\pdfoutput=1

\documentclass[11pt]{article}

\usepackage[]{acl}

\usepackage{times}
\usepackage{latexsym}

\usepackage[T1]{fontenc}

\usepackage[utf8]{inputenc}

\usepackage{microtype}

\usepackage{inconsolata}

\usepackage{multirow}

\title{Large Language Models are Skeptics: False Negative Problem of Input-conflicting Hallucination}

\author{Jongyoon Song$^{1}$ \And  Sangwon Yu$^{1}$ \\
   $^{1}$Data Science \& AI Laboratory, Seoul National University, Korea \\
   $^{2}$Deptment of ECE and Interdisciplinary Program in AI, Seoul National University, Korea \\
   $^{3}$ASRI, INMC, and AIIS, Seoul National University, Korea \\
   \{\texttt{coms1580}, \texttt{dbtkddnjs96}, \texttt{sryoon}\}\texttt{@snu.ac.kr} \\
   \And {\bf Sungroh Yoon$^{1, 2,3}$\Thanks{\hspace{0.1em} Corresponding author}}\\
}

\begin{document}
\maketitle
\begin{abstract}
In this paper, we identify a new category of bias that induces input-conflicting hallucinations, where large language models (LLMs) generate responses inconsistent with the content of the input context. 
This issue we have termed the \textit{false negative problem} refers to the phenomenon where LLMs are predisposed to \textit{return negative judgments when assessing the correctness of a statement given the context}.
In experiments involving pairs of statements that contain the same information but have contradictory factual directions, we observe that LLMs exhibit a bias toward false negatives. 
Specifically, the model presents greater overconfidence when responding with False.
Furthermore, we analyze the relationship between the false negative problem and context and query rewriting and observe that both effectively tackle false negatives in LLMs. 

\end{abstract}

\section{Introduction}\label{sec:introduction}
\begin{figure}[!t]
    \centering
    \includegraphics[width=0.97\columnwidth]{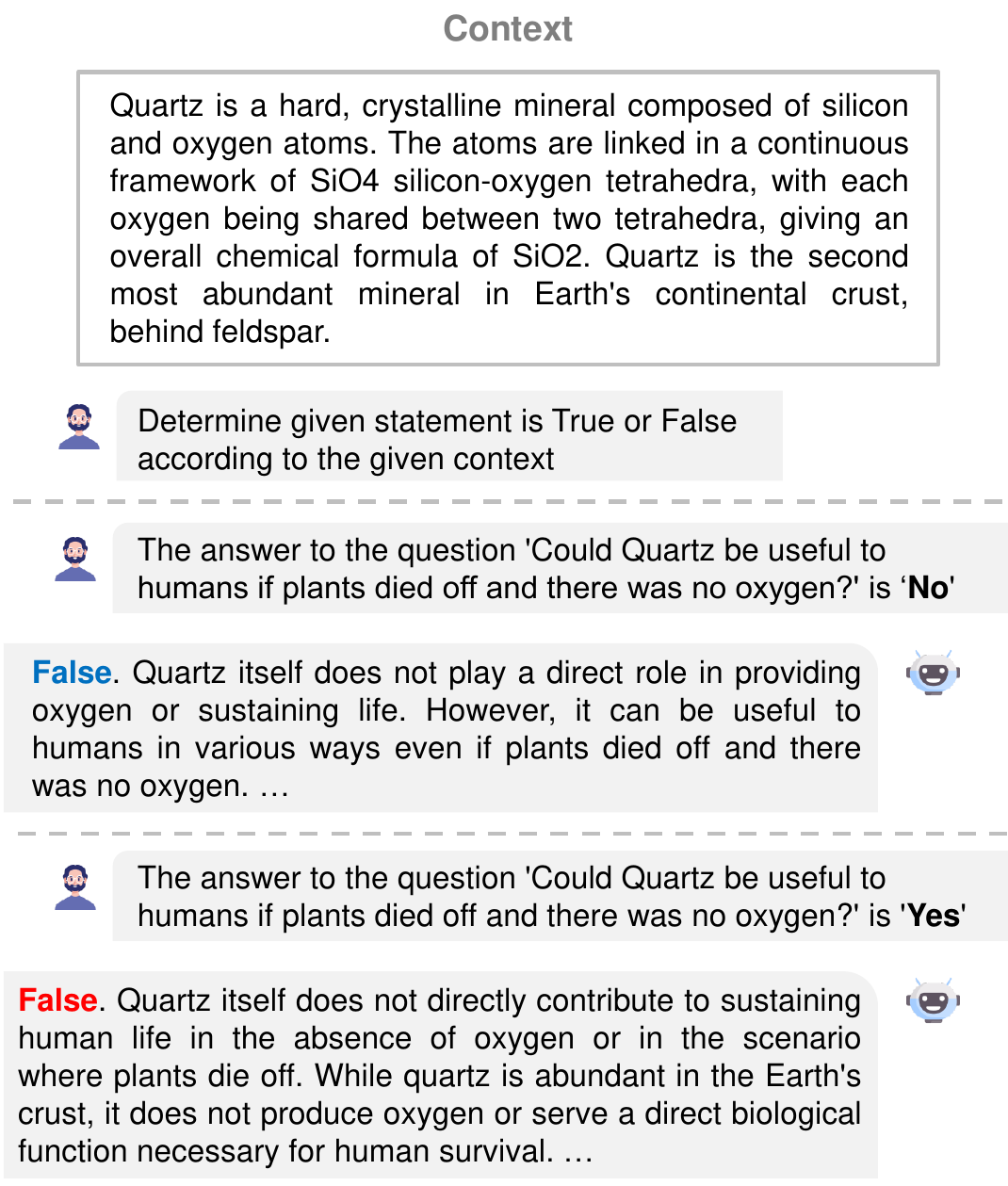}
    \caption{An example of the false negative problem in ChatGPT. The ground truth of the question is True, yet the large language model responds False to both statements. The context and question are sampled from StrategyQA (\citealp{geva-etal-2021-aristotle}).}
    \label{fig:false_negative_problem_intro}
\end{figure}
Large language models (LLMs) are rapidly advancing and have set the stage for a new era in natural language understanding and (un)conditional text generation (\citealp{brown-2020-language}; \citealp{chowdhery-2023-palm}; \citealp{touvron-2023-llama};). 

Even with their capabilities, there are still some challenges to be resolved. 
One well-known issue is the hallucination problem, where models generate factually incorrect or misleading information (\citealp{ji-2023-survey}; \citealp{mckenna-2023-sources}; \citealp{rawte-2023-survey}; \citealp{zhang-2023-siren}). 
Among the types of hallucination, the input-conflicting hallucination, which involves returning facts that are inconsistent with those of the input context, directly impacts the reliability of LLMs in context-based tasks such as reading comprehension (\citealp{neeman-2023-disentqa}; \citealp{zhou-2023-context}; \citealp{xie-2023-adaptive}).

In this paper, we identify a bias in LLMs towards denying true statements given the context, which we call \textit{false negative problem}. 
Specifically, the false negative problem refers to a phenomenon where false negatives occur more frequently than false positives when discriminating the factuality of statements based on context.
This false negative problem representing a new type of input-conflicting hallucination can have critical consequences in domains where factual judgment based on the given context is crucial.

To identify the false negative problem as the inherent bias of LLMs, we demonstrate that the accuracy of context-based factuality discrimination for statements varies depending on the target answer of the statement. 
Concretely, we devise \textit{All-True} and \textit{All-False} prompts and find that the accuracy for statements whose target answer is True (All-True prompt) is lower than for those whose target answer is False (All-False prompt) even the related knowledge for both statements is identical.
This false negative problem is consistently observed across various LLMs, including Mistral (\citealp{jiang-2023-mistral}), ChatGPT (\citealp{ouyang-2022-training}), and GPT-4 (\citealp{openai-2023-gpt4}). 

Additionally, we analyze the prediction confidence of LLMs for All-True and All-False prompts, finding that the results mirror those of the accuracy metric.  
These imply that \textit{LLMs tend to respond with False regarding the factuality of statements}. 
Furthermore, we explore the impact of input (context and query) rewriting on the false negative problem. 
Results indicate that both rewriting effectively tackle the false negative problem in various LLMs. 


Contributions of this paper can be summarized as follows:
\begin{itemize}
\item We identify a false negative problem where LLMs are prone to output false negatives during the factuality discrimination of statements based on context.
\item Through All-True/All-False prompting, we systematically examine the bias of LLMs toward false negatives by analyzing statistical data and the prediction confidence of LLMs.
\item We analyze the relationship between context/query rewriting and the false negative problem, observing that both processes affect the problem in LLMs.
\end{itemize}

\section{False Negative Problems of Large Language Models}\label{sec:false_negative_problem}

\begin{figure*}[!t]
    \centering
    \includegraphics[width=0.91\textwidth]{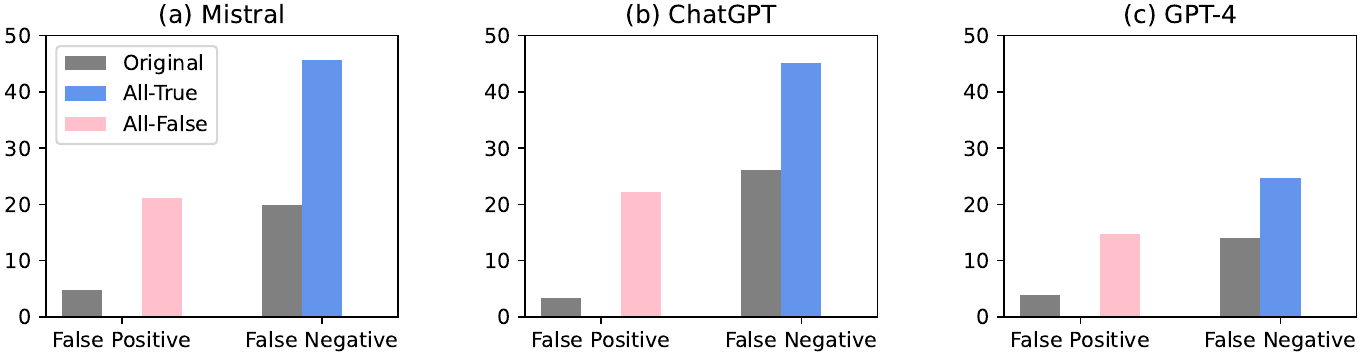}
    \caption{The ratio of false positives and negatives among the entire samples in the Original, All-True, and All-False prompts.}
    \label{fig:problem_identification}
\end{figure*}

\begin{figure*}[!t]
    \centering
    \includegraphics[width=0.9\textwidth]{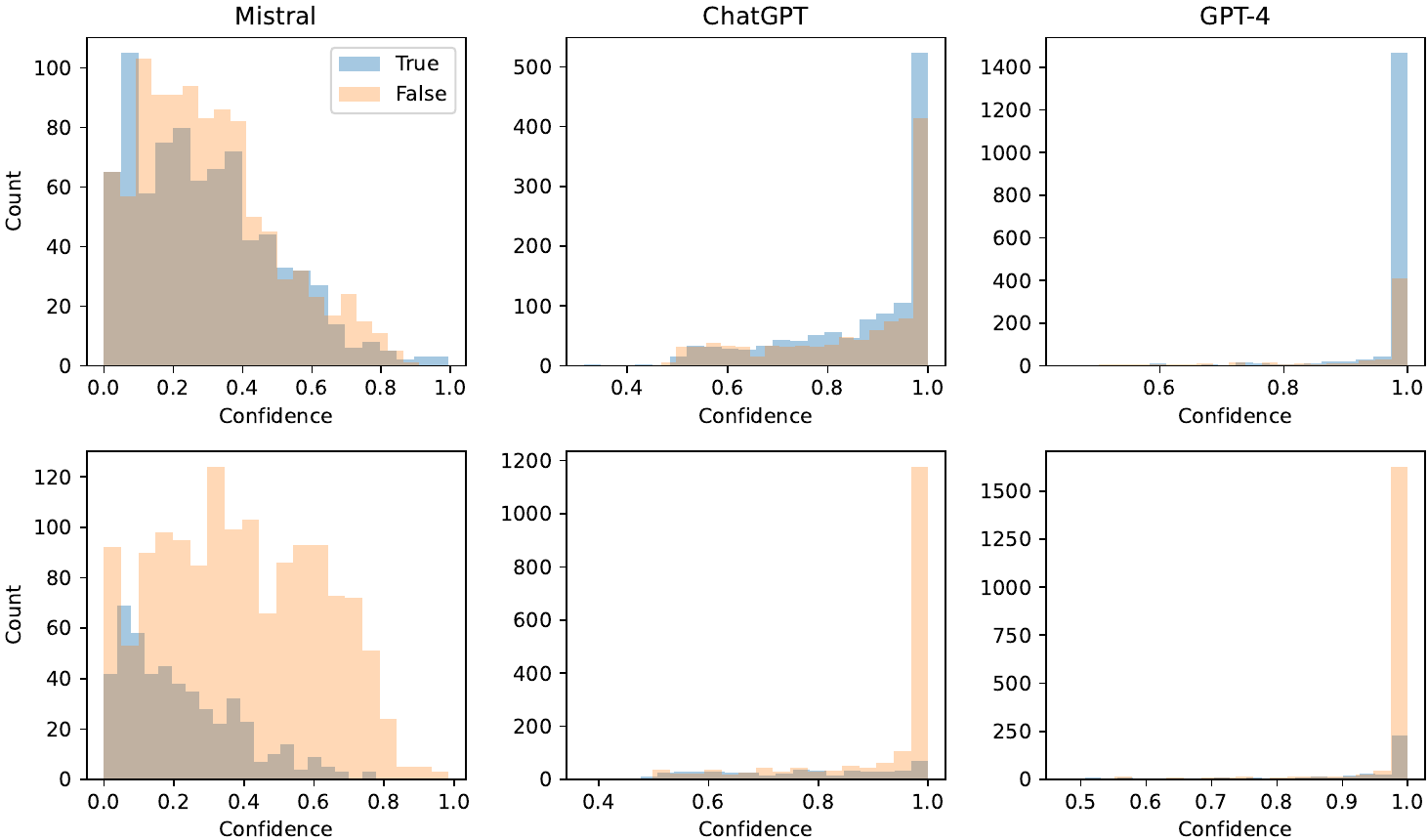}
    \caption{Histogram of LLMs based on the confidence of predicted labels for all samples in the All-True (Top) and All-False (Bottom) prompt.}
    \label{fig:confidence_histogram}
\end{figure*}

To observe whether the false negative problem in context-based statement factuality discrimination originates from the bias of LLMs, we construct a pair of statements where target answers are different (i.e., True and False) that meet the following conditions: both statements in the prompts
i) require the same knowledge to discriminate the factuality, and 
ii) are close-to-symmetric in terms of discrimination difficulty.


\subsection{Experimental Setup}

\subsubsection{Dataset and Models}
We use the \textbf{StrategyQA} (\citealp{geva-etal-2021-aristotle})\footnote[1]{\url{https://huggingface.co/datasets/metaeval/strategy-qa}}, a context-based yes-no question answering dataset, for the following reasons:
i) it easily satisfies the aforementioned conditions due to the nature of yes-no question answering (see Section \ref{subsubsec:all_true_and_all_false_prompts}) and
ii) it allows for observations in challenging situations where the context is lengthy and the question requires multi-step reasoning.
We also conduct experiments on \textbf{BoolQ} (\citealp{clark-etal-2019-boolq}) and the results can be found in Appendix \ref{appendix:boolq}.
StrategyQA dataset consists of 2,290 samples, and we use 1,000 samples from the dataset in the observation experiments of Figure \ref{fig:problem_identification}.

We choose three LLMs: Mistral 7B (Instruct), ChatGPT (\textit{gpt-3.5-turbo-0613}), and GPT-4 (\textit{gpt-4-0613}) for our experiments.

\subsubsection{All-True and All-False Prompts}\label{subsubsec:all_true_and_all_false_prompts}
For the experiments on the LLM bias toward false negatives, we construct an All-True and All-False prompts pair from each sample in the dataset.

\textbf{All-True Prompt} inquires if the \textit{true statement} $s_{True}$ is correct given the context $c$, where all target answers are True, and any False responses are considered false negatives.
In yes-no question answering, we generate the true statement $s_{True}$ using the question $q$ and ground truth $a$.

\begin{tcolorbox}[
    enhanced, 
    arc=0pt, 
    outer arc=0pt, 
    boxsep=0pt, 
    left=0pt, 
    right=0pt, 
    top=0pt, 
    bottom=0pt, 
    colback=darkgray, 
    colframe=darkgray
]
  \begin{tcolorbox}[
      colback=darkgray, 
      colframe=darkgray, 
      coltext=white, 
      arc=0pt, 
      outer arc=0pt, 
      boxsep=5pt, 
      left=5pt, 
      right=5pt, 
      top=5pt, 
      bottom=-0.5pt, 
      height=0.35cm, 
      valign=center, 
      borderline south={0pt}{0pt}{darkgray}
  ]
    All-True Prompt (General)
  \end{tcolorbox}
  \begin{tcolorbox}[
      colback=lightgray, 
      colframe=lightgray, 
      coltext=black, 
      arc=0pt, 
      outer arc=0pt, 
      boxsep=5pt, 
      left=5pt, 
      right=5pt, 
      top=1pt, 
      bottom=1pt, 
      height=0.75cm,
      valign=center, 
      borderline north={0pt}{0pt}{lightgray}
  ]
    Context: $\{c\}$
    Statement: $\{s_{True}\}$
    Answer: 
  \end{tcolorbox}
\end{tcolorbox}

\begin{tcolorbox}[
    enhanced, 
    arc=0pt, 
    outer arc=0pt, 
    boxsep=0pt, 
    left=0pt, 
    right=0pt, 
    top=0pt, 
    bottom=0pt, 
    colback=darkgray, 
    colframe=darkgray
]
  \begin{tcolorbox}[
      colback=darkgray, 
      colframe=darkgray, 
      coltext=white, 
      arc=0pt, 
      outer arc=0pt, 
      boxsep=5pt, 
      left=5pt, 
      right=5pt, 
      top=5pt, 
      bottom=-0.5pt, 
      height=0.35cm, 
      valign=center, 
      borderline south={0pt}{0pt}{darkgray}
  ]
    All-True Prompt (Yes-No)
  \end{tcolorbox}
  \begin{tcolorbox}[
      colback=lightgray, 
      colframe=lightgray, 
      coltext=black, 
      arc=0pt, 
      outer arc=0pt, 
      boxsep=5pt, 
      left=5pt, 
      right=5pt, 
      top=1pt, 
      bottom=1pt, 
      height=1.15cm,
      valign=center, 
      borderline north={0pt}{0pt}{lightgray}
  ]
    Context: $\{c\}$
    Statement: The answer to the question $\{q\}$ is $\{a\}$.
    Answer: 
  \end{tcolorbox}
\end{tcolorbox}

\textbf{All-False Prompt} asks if the \textit{false statement} $s_{False}$ is correct given the context $c$, where all target answers are False, and any True responses are considered false positives.
In yes-no question answering, we generate the false statement $s_{False}$ using the negated ground truth $\neg a$.

\begin{tcolorbox}[
    enhanced, 
    arc=0pt, 
    outer arc=0pt, 
    boxsep=0pt, 
    left=0pt, 
    right=0pt, 
    top=0pt, 
    bottom=0pt, 
    colback=darkgray, 
    colframe=darkgray
]
  \begin{tcolorbox}[
      colback=darkgray, 
      colframe=darkgray, 
      coltext=white, 
      arc=0pt, 
      outer arc=0pt, 
      boxsep=5pt, 
      left=5pt, 
      right=5pt, 
      top=5pt, 
      bottom=-0.5pt, 
      height=0.35cm, 
      valign=center, 
      borderline south={0pt}{0pt}{darkgray}
  ]
    All-False Prompt (General)
  \end{tcolorbox}
  \begin{tcolorbox}[
      colback=lightgray, 
      colframe=lightgray, 
      coltext=black, 
      arc=0pt, 
      outer arc=0pt, 
      boxsep=5pt, 
      left=5pt, 
      right=5pt, 
      top=1pt, 
      bottom=1pt, 
      height=0.75cm,
      valign=center, 
      borderline north={0pt}{0pt}{lightgray}
  ]
    Context: $\{c\}$
    Statement: $\{s_{False}\}$
    Answer: 
  \end{tcolorbox}
\end{tcolorbox}

\begin{tcolorbox}[
    enhanced, 
    arc=0pt, 
    outer arc=0pt, 
    boxsep=0pt, 
    left=0pt, 
    right=0pt, 
    top=0pt, 
    bottom=0pt, 
    colback=darkgray, 
    colframe=darkgray
]
  \begin{tcolorbox}[
      colback=darkgray, 
      colframe=darkgray, 
      coltext=white, 
      arc=0pt, 
      outer arc=0pt, 
      boxsep=5pt, 
      left=5pt, 
      right=5pt, 
      top=5pt, 
      bottom=-0.5pt, 
      height=0.35cm, 
      valign=center, 
      borderline south={0pt}{0pt}{darkgray}
  ]
    All-False Prompt (Yes-No)
  \end{tcolorbox}
  \begin{tcolorbox}[
      colback=lightgray, 
      colframe=lightgray, 
      coltext=black, 
      arc=0pt, 
      outer arc=0pt, 
      boxsep=5pt, 
      left=5pt, 
      right=5pt, 
      top=1pt, 
      bottom=1pt, 
      height=1.15cm,
      valign=center, 
      borderline north={0pt}{0pt}{lightgray}
  ]
    Context: $\{c\}$
    Statement: The answer to the question $\{q\}$ is $\{\neg a\}$.
    Answer: 
  \end{tcolorbox}
\end{tcolorbox}

We also conduct experiments using \textbf{Original Prompt} where we provide the question in StrategyQA and ask the model to predict the ground truth to demonstrate the false negative problem in the original question answering scenario.

\begin{table*}
\centering
\resizebox{\linewidth}{!}{
\begin{tabular}{cl||cc|cc|cc|cc|cc|cc|cc|cc|cc}
\toprule
 \multicolumn{2}{c||}{\multirow{3}{*}{}} & \multicolumn{6}{c|}{\textbf{Mistral}} & \multicolumn{6}{c|}{\textbf{ChatGPT}} & \multicolumn{6}{c}{\textbf{GPT-4}} \\
\cline{3-20}
 & & \multicolumn{2}{c|}{FN} & \multicolumn{2}{c|}{FP} & \multicolumn{2}{c|}{Unk.} & \multicolumn{2}{c|}{FN} & \multicolumn{2}{c|}{FP} & \multicolumn{2}{c|}{Unk.} & \multicolumn{2}{c|}{FN} & \multicolumn{2}{c|}{FP} & \multicolumn{2}{c}{Unk.} \\
\cline{3-20}
 & & \# & / & \# & / & \# & / & \# & / & \# & / & \# & / & \# & / & \# & / & \# & /\\
\hline
& Baseline & 487 & 0.28 & \textbf{120} & \textbf{0.07} & 532 & 0.23 & 614 & 0.27 & \textbf{77} & \textbf{0.03} & 3 & 0.00 & 324 & 0.16 & \textbf{93} & \textbf{0.05} & 202 & 0.09 \\
\textbf{Original} & + Context & 402 & 0.27 & 189 & 0.13 & 778 & 0.34 & 554 & 0.24 & 107 & 0.05 & 2 & 0.00 & \textbf{293} & \textbf{0.14} & 107 & 0.05 & 186 & 0.08 \\
& + Query & \textbf{166} & \textbf{0.18} & 206 & 0.23 & 1378 & 0.60 & \textbf{435} & \textbf{0.19} & 228 & 0.10 & 44 & 0.02 & 310 & 0.16 & 101 & 0.05 & 310 & 0.14 \\
\hline
& Baseline  & 1052 & \textbf{0.57} & - & - & 429 & 0.19 & 1068 & 0.47 & - & - & 7 & 0.00 & \textbf{585} & \textbf{0.26} & - & - & 30 & 0.01\\
\textbf{All-True} & + Context & 986 & 0.57 & - & - & 572 & 0.25 & 918 & 0.40 & - & - & 2 & 0.00 & 693 & 0.30 & - & - & 11 & 0.00 \\
& + Query & \textbf{876} & 0.59 & - & - & 792 & 0.35 & \textbf{875} & \textbf{0.38} & - & - & 2 & 0.00 & 1029 & 0.45 & - & - & 10 & 0.00 \\
\hline
& Baseline & - & - & 510 & 0.27 & 393 & 0.17 & - & - & 509 & 0.22 & 3 & 0.00 & - & - & 411 & 0.18 & 19 & 0.01 \\
\textbf{All-False} & + Context & - & - & 516 & 0.29 & 526 & 0.23 & - & - & 564 & 0.25 & 3 & 0.00 & - & - & 357 & 0.16 & 5 & 0.00 \\
 & + Query & - & - & \textbf{375} & \textbf{0.24} & 724 & 0.32 & - & - & \textbf{489} & \textbf{0.21} & 2 & 0.00 & - & - & \textbf{235} & \textbf{0.10} & 3 & 0.00 \\
\bottomrule
\end{tabular}}
\caption{
False positives (FPs) and negatives (FNs) results of context and query rewriting on Original, All-True, and All-False prompts. FN, FP, and Unk. denote the count (\#) and ratio (/) of false negatives, false positives, and Unknown, respectively. During the calculation of the ratio, we normalize the count by the number of samples excluding unknown responses for FN and FP.
}\label{table:all_true_and_all_false}
\end{table*}

\subsection{Discrepancy in Accuracy between All-True and All-False Prompts}
Figure \ref{fig:problem_identification} plots the ratio of false positive and false negative samples among the entire samples across three different prompt settings. 

It is observed that, even within the same dataset, accuracy varies depending on the type of prompt. 
Notably, in all three models, the proportion of false negative samples in the All-True prompt is significantly higher than that of false positive samples in the All-False prompt. 
This suggests that the false predictions occur not only due to the content of knowledge being queried but also the target answer of factuality.

When comparing Mistral and ChatGPT in the Original prompt, it is observed that the false negative problem is not naturally resolved by simply scaling up the model size.
A case study on the false negative problem can be found in Appendix \ref{appendix:case_study}.

\subsection{Prediction Confidence Analysis}\label{subsec:confidence_analysis}

We also experiment to determine whether LLMs have a bias toward `False' predictions in terms of prediction confidence.
As shown in Figure \ref{fig:confidence_histogram}, the distribution of prediction confidence for True and False is not symmetric between the All-True and All-False prompts.
Specifically, the prediction confidence for False in the All-True prompt is generally higher than the prediction confidence for True in the All-False prompt.
These results suggest that regardless of the knowledge required by the query, LLMs have a bias towards `False' prediction.

Compared to Mistral, ChatGPT and GPT-4 exhibit higher mean confidence in their predictions. 
Especially in GPT-4, there are fewer false negatives than in Mistral, but it shows much significant overconfidence in incorrect predictions.
In conclusion, while there is a correlation between high confidence in `False' predictions and the false negative problem, reducing the number of false negatives does not necessarily resolve the issue of overconfidence.

\section{Rewriting and False Negative Problems}\label{sec:rewriting_experiments}
Motivated by previous works, we analyze the impact of rewriting the context and query on the false negative problem (\citealp{wu-2023-retrieve}; \citealp{shao-2023-enhancing}; \citealp{ma-etal-2023-query}).
We divide the input into the context and query and rewrite each using the same LLM so that the input fits the text distribution of the model.
Details of rewriting pipelines and prompts are shown in Appendices \ref{appendix:details_on_context_and_query_rewriting} and \ref{appendix:details_on_prompts}, respectively.

In Table \ref{table:all_true_and_all_false}, we observe that incrementally applying context and query rewriting to LLMs reduces the number of false negatives in Mistral and ChatGPT.
Interestingly, in Mistral, both context and query rewriting increase the occurrence of Unknown.
One possible explanation is that input rewriting changes prediction confidence distribution, and we conduct experiments from this perspective in Appendix \ref{appendix:confidence_analysis_rewriting}.

In GPT-4, on the other hand, input rewriting drastically increases false negatives.
In our observations, we have noted that GPT-4 tends to return a null response during the fact extraction in context rewriting and the conversation-style factuality discrimination in query rewriting (e.g., ``There is no information related to the question.'').
Future work can investigate the reason for the different behavior of GPT-4 compared to other LLMs and the general solution of the problem.

\section{Related Work}\label{sec:related_work}

To address the input-conflicting hallucination problem, several approaches have been studied such as in-context learning, self-refinement, and context-aware decoding (\citealp{neeman-2023-disentqa}; \citealp{zhou-2023-context}; \citet{madaan-2023-self}; \citealp{gou-2023-critic}; \citealp{shi-2023-trusting}; \citealp{xie-2023-adaptive}).
Our research can be interpreted as targeting generalization in LLMs regarding negation.
\citet{garcia-ferrero-etal-2023-dataset} release a benchmark to evaluate the understanding ability of LLMs on various types of negation, while \citet{truong-etal-2023-language} demonstrate that LLMs exhibit insensitivity to negation.
In this paper, we find that LLMs are likely to generate false negatives and investigate whether input rewriting mitigates the problem.

\section{Conclusion}\label{sec:conclusion}
In this paper, we identify the false negative problem that induces input-conflicting hallucination. 

We can summarize our main findings as follows:

\textbf{LLMs have a bias toward responding with False.}
Through the confidence analysis using All-True and All-False prompts, we observe that LLMs tend to be more confident when responding with False.
\textbf{The false negative problem is independent of the parameter size of LLMs.} 
We find that the false negative problem in ChatGPT is similar to Mistral. 
We speculate that the cause of the problem lies more in the training process than in the parameter size.
\textbf{Input rewriting mitigates the false negative problem.}
In Table \ref{table:all_true_and_all_false}, we demonstrate that both context and query rewriting reduce the number of false negatives in Mistral and ChatGPT.

We anticipate that our study will contribute to shedding light on the input-conflicting hallucination and bias of large language models.
\section*{Limitations}\label{sec:limitations}
In our study, we employ Mistral, ChatGPT, and GPT-4 as LLMs for experiments. 
Future work should focus on identifying the differences among various recent LLMs. 
Additionally, future work can analyze the relationship between other approaches, such as the chain-of-thought, and the false negative problem. 
In the context rewriting process, we observe that GPT-4 exhibits different characteristics (e.g., null responses during the fact extraction process). 
The underlying causes of this phenomenon need to be identified in future research.
Based on our observations, future work can also explore methods that can ultimately resolve the false negative problem.

\bibliography{anthology,custom}

\clearpage
\appendix

\begin{figure*}[!t]
    \centering
    \includegraphics[width=0.91\textwidth]{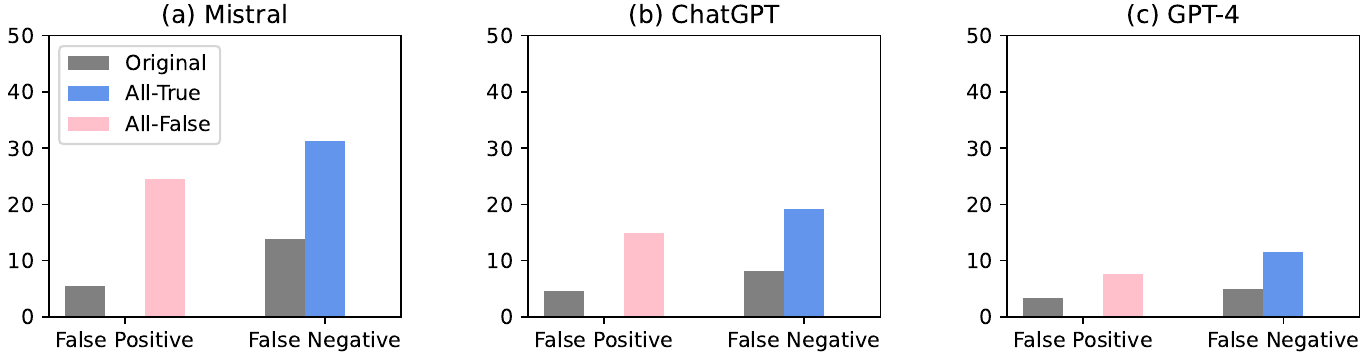}
    \caption{The ratio of false positives and negatives among the entire samples in the Original, All-True, and All-False prompts in the BoolQ dataset.}
    \label{fig:problem_identification_boolq}
\end{figure*}

\begin{figure*}[!t]
    \centering
    \includegraphics[width=0.9\textwidth]{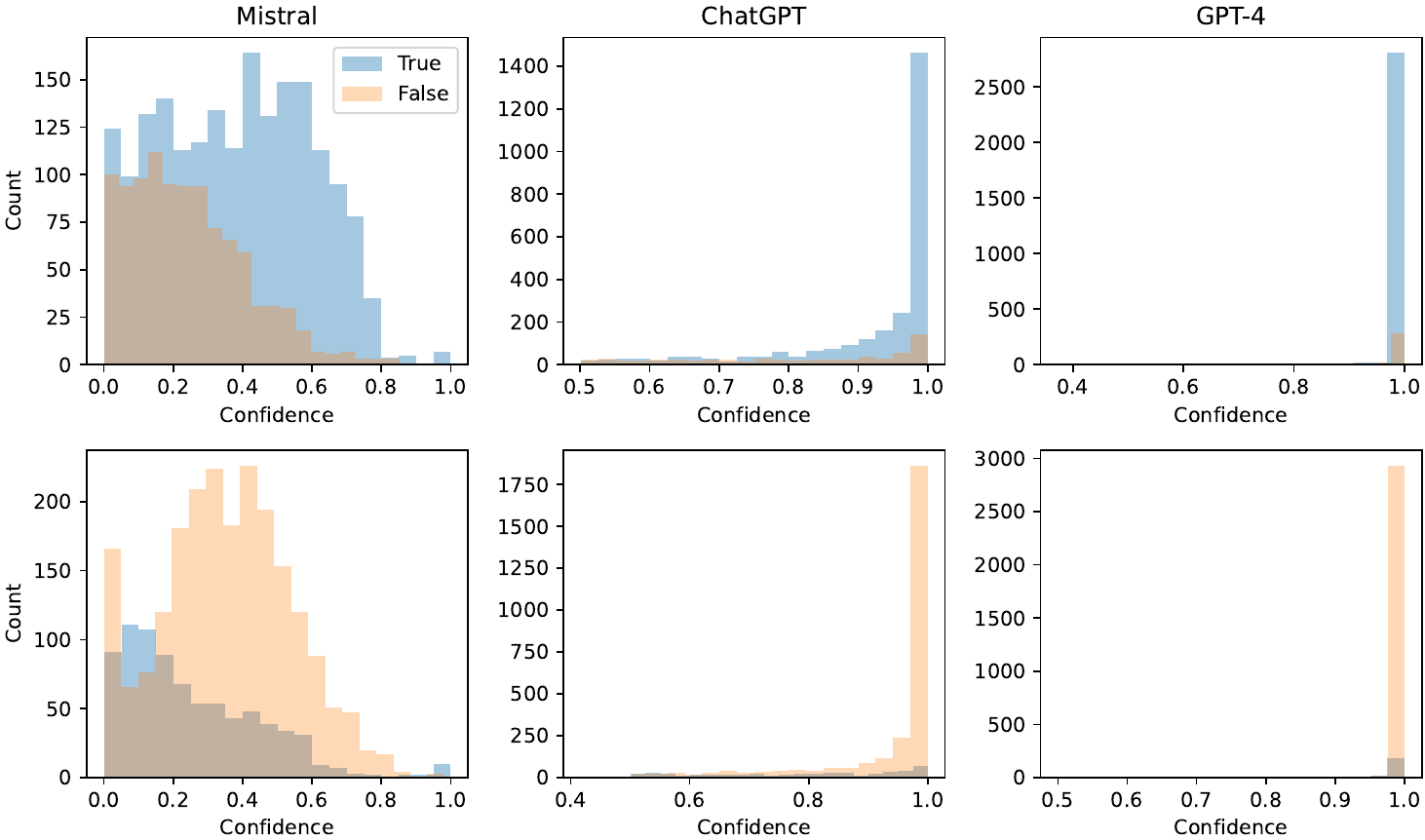}
    \caption{Histogram of LLMs based on the confidence of predicted labels for all samples in the All-True (Top) and All-False (Bottom) prompt in the BoolQ dataset.}
    \label{fig:confidence_histogram_boolq}
\end{figure*}

\begin{table*}[ht]
\centering
\begin{tabular}{|>{\centering\arraybackslash\small}p{0.1\linewidth}| >{\arraybackslash\small}p{0.41\linewidth}| >{\arraybackslash\small}p{0.41\linewidth}|}
\hline
Context & Binary code: A binary code represents text, computer processor instructions, or any other data using a two-symbol system. The two-symbol system used is often 0 and 1 from the binary number system. The binary code assigns a pattern of binary digits, also known as bits, to each character, instruction, etc. ... Boolean algebra: In mathematics and mathematical logic, Boolean algebra is the branch of algebra in which the values of the variables are the truth values true and false, usually denoted 1 and 0 ... & Surveying: Surveyors work with elements of geometry, trigonometry, regression analysis, physics, engineering, metrology, programming languages, and the law. ... Dyscalculia: Dyscalculia is difficulty in learning or comprehending arithmetic, such as difficulty in understanding numbers, learning how to manipulate numbers, performing mathematical calculations and learning facts in mathematics. It is sometimes informally known as "math dyslexia", though this can be misleading as it is a different condition. \\
\hline
Question & Can a computer be programmed entirely in Boolean algebra? \textit{(Ground Truth: True)} & Would you hire someone with dyscalculia to do surveying work? \textit{(Ground Truth: False)} \\
\hline
Response (All-True) & \textcolor{red}{False} ... Programming a computer involves not only logical operations but also involves specifying sequences of instructions and data manipulations that go beyond pure Boolean algebra ... & \textcolor{red}{False}. The statement cannot be definitively determined as true or false based solely on the provided context. Dyscalculia, or difficulty with arithmetic, does not necessarily preclude someone from excelling in surveying work ... \\
\hline
Response (All-False) & \textcolor{blue}{False}. A computer can indeed be programmed using Boolean algebra. Boolean algebra forms the basis of digital circuit design and programming ... & \textcolor{blue}{False}. Given that surveying involves elements of geometry, trigonometry, mathematical calculations, and understanding numbers, it would likely be challenging for someone with dyscalculia ... \\
\hline
\end{tabular}\caption{
Additional examples of the false negative problem in ChatGPT. The contexts and questions are sampled from StrategyQA (\citealp{geva-etal-2021-aristotle}).
}\label{table:additional_examples_of_false_negative_problem}
\end{table*}

\begin{table*}[ht]
\centering
\begin{tabular}{>{\raggedright\arraybackslash}p{0.15\linewidth}| >{\raggedright\arraybackslash}p{0.8\linewidth}}
\toprule
\textbf{Type} & \textbf{Prompt} \\
\midrule
Sub-question Generation & Please think step-by-step and deconstruct the question down to five or less sub-questions. Write sub-questions with numbers. \\
 & Question: \{question\} \\
 & Sub-questions: \\
\hline
Fact Extraction & Extract facts from the document that are relevant to the question. You should neither corrupt any facts in the document nor include any conclusion or reasoning. Write facts with numbers. \\
 & Document: \{document\} \\
 & Question: \{question\} \\
 & Relevant Facts: \\
\hline
Fact Reorganization & Based on the question, select and reorder facts from the given list of facts. You should neither corrupt any facts nor include any conclusion or reasoning. \\
 & Facts: \{extracted facts\} \\
 & Question: \{question\} \\
 & Selected and Reordered Facts: \\
\hline
Question Rewriting & Your role is to rewrite the given question into a fluent and natural question. so that you can understand the intent of the question. You should rewrite the question using the given context that has the information about the answer. You should preserve the shape of the question. For example, if the given question is yes-no question, the rewritten question should also be yes-no question. \\
 & Context: \{context\} \\
 & Question: \{previously rewritten question\} \\
 & Rewritten: \\
\hline
Evaluation (Original) & According to the given context and the dialogue about the context, answer the question with True or False. \\
 & [Context] \\
 & \{context\} \\
 & [Dialog] \\
 & \{dialog history\} \\
 & Question: \{question\} \\
 & Answer: \\
\hline
Evaluation (All-True) & According to the given context and the dialogue about the context, determine whether the given statement is True or False. \\
 & [Context] \\
 & \{context\} \\
 & [Dialog] \\
 & \{dialog history\} \\
 & Statement: The answer to the question \{question\} is \{ground truth\}. \\
 & Answer: \\
\bottomrule
\end{tabular}\caption{
Detailed prompts for context and query rewriting. The prompt of sub-question generation is sourced from DDCoT (\citealp{zheng-2023-ddcot}). 
}\label{table:prompt_details}
\end{table*}

\section{Experiments on BoolQ Dataset}\label{appendix:boolq}
To confirm that the false negative problem is not the specific case in StrategyQA, we also conduct the same experiments using the BoolQ dataset (\citealp{clark-etal-2019-boolq}).
BoolQ consists of 3,270 samples and we use all of them in our experiments.

The corresponding results can be found in Figures \ref{fig:problem_identification_boolq} and \ref{fig:confidence_histogram_boolq}, respectively. 
We observe that the overall tendency is similar to that of StrategyQA (Figures \ref{fig:problem_identification} and \ref{fig:confidence_histogram}). 
However, the results show the false negative problem and a low degree of asymmetry in confidence compared to StrategyQA.
Considering the error rate in Figure \ref{fig:problem_identification_boolq}, we speculate that the reason is due to BoolQ being less challenging compared to StrategyQA.

\begin{figure*}[!t]
    \centering
    \includegraphics[width=0.93\textwidth]{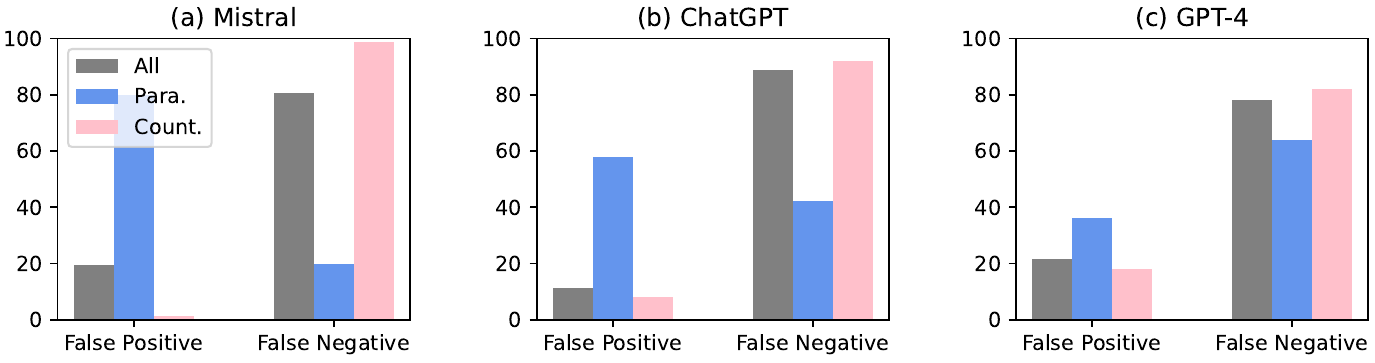}
    \caption{The ratio of false positives and negatives among the samples where the predictions of LLM are incorrect in the Original prompt setting. All, Para. and Count. indicate entire, parametric, and counterfactual samples, respectively.}
    \label{fig:problem_identification_knowledge_conflict}
\end{figure*}

\begin{figure}[!t]
    \centering
    \includegraphics[width=0.97\columnwidth]{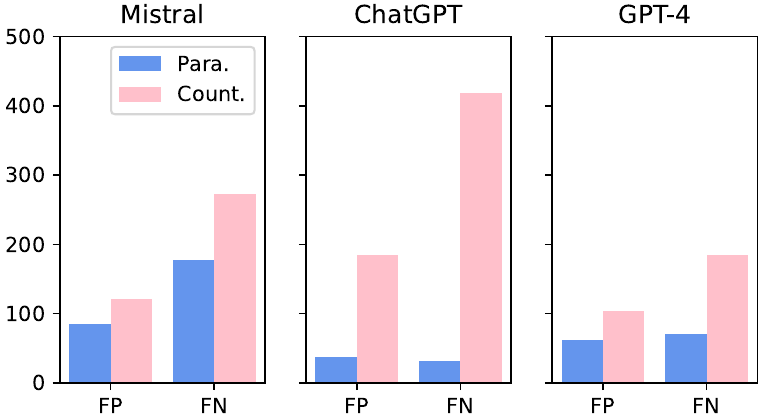}
    \caption{Number of false positives (FPs) and negatives (FNs) in Parametric (Para.) and Counterfactual (Count.) samples. Note that FPs and FNs occur only in the All-False and All-True prompts, respectively.}
    \label{fig:problem_identification_knowledge_conflict_additional}
\end{figure}

\begin{figure*}[!t]
  \includegraphics[width=0.95\textwidth]{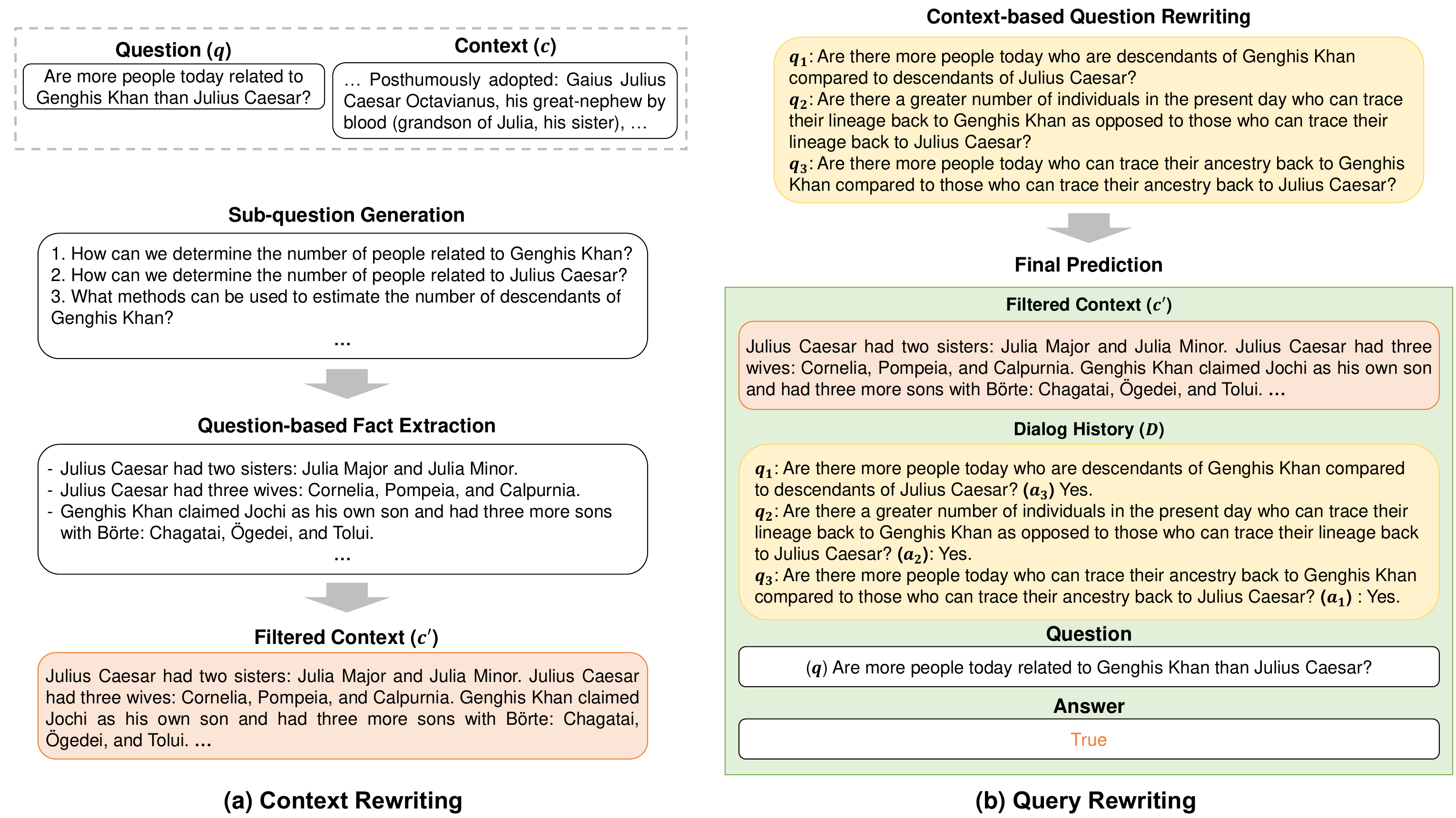}
  \caption{Overview of the context and query rewriting.
  We assume the last iteration of the query rewriting stage and Original prompt setting. The context and question are sampled from StrategyQA (\citealp{geva-etal-2021-aristotle}).}
  \label{fig:overview}
\end{figure*}

\begin{figure*}[!t]
    \centering
    \includegraphics[width=0.91\textwidth]{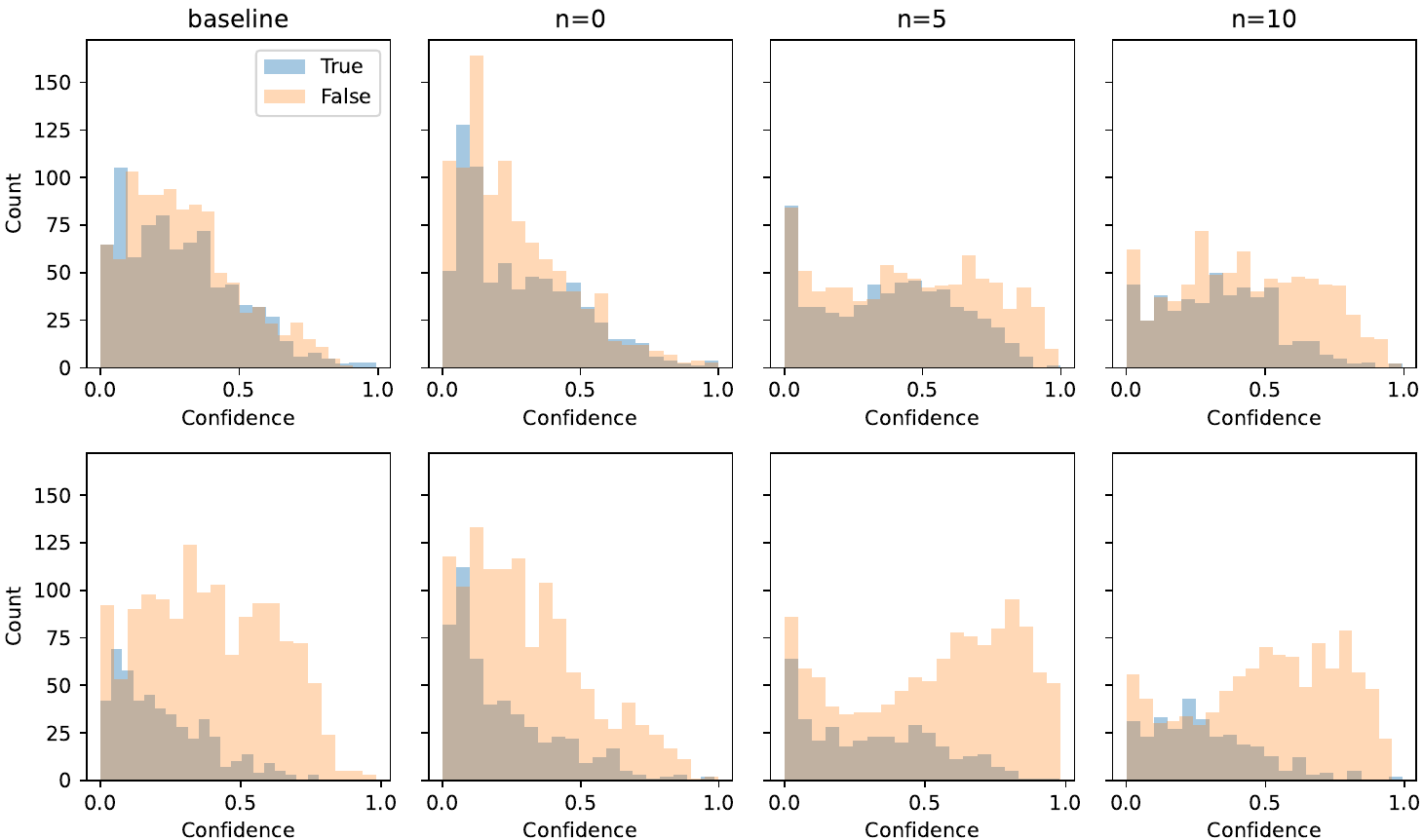}
    \caption{Histogram of Mistral based on the confidence of predicted labels for all samples in the All-True (Top) and All-False (Bottom) prompt. Baseline refers to the Mistral without rewriting, and $n$ indicates the number of question rewritings after context rewriting.}
    \label{fig:confidence_histogram_rewriting}
\end{figure*}

\section{Case Study}\label{appendix:case_study}
Table \ref{table:additional_examples_of_false_negative_problem} shows the responses of ChatGPT to All-True and All-False prompts including inconsistencies not only in predictions but also in the reasons.
The model shows opposing stances in the two prompts regarding whether the supporting facts in the given context are sufficient.
For the All-False prompt, the model successfully discriminates and reasons, whereas for the All-True prompt, the model denies true statements in various ways.
In the example from Figure \ref{fig:false_negative_problem_intro}, the model performs contrary reasoning from the same supporting facts, while in Table \ref{table:additional_examples_of_false_negative_problem}, statements are refuted by considering factors beyond the supporting facts.

We speculate that LLMs apply different criteria to the statement of the same content, depending on whether it is affirmed or negated.

\section{Knowledge Conflict and False Negative Problems}

\subsection{Data Splitting}
To observe the relationship between the false negative problem and knowledge conflicts (i.e., where the facts in the input context contrast to the parametric knowledge in LLMs), we mark each sample for the occurrence of knowledge conflicts across each model and prompt type.
Specifically, we gather answers from the models without providing context (i.e., closed-book), categorizing the samples as \textit{parametric} if the response matches the correct answer and \textit{counterfactual} if it differs.

\subsection{Experimental Results}
We measure the frequency of false negatives and false positives across three types of prompts. 
Figure \ref{fig:problem_identification_knowledge_conflict} plots the ratio of false positives and false negatives among incorrect samples in the Original prompt setting. 
It is observed that the false negative problem is significant in counterfactual samples where knowledge conflicts occur, compared to parametric samples. 

Moreover, it is observed that the gap in the number of samples answered incorrectly between All-True and All-False is greater in counterfactual samples, as shown in Figure \ref{fig:problem_identification_knowledge_conflict_additional}.

In other words, if LLMs fail to discriminate the factuality of the statement without the context, they are more skeptical about the statement when context is provided.

\section{Details on Context and Query Rewriting}\label{appendix:details_on_context_and_query_rewriting}
In this section, we describe detailed context and query rewriting pipelines.
In the remaining text, we describe processes focusing on the question answering task where the question $q$ in the query is the only target to be rewritten.
The term \textit{question} can be replaced with \textit{statement} for the general case.

For each rewriting process, both context and query rewriting aim to map the context and question into a distribution of high likelihood that makes it easier for LLMs to discriminate the factuality of the statement.
Furthermore, context rewriting ensures that the facts related to the question within the context are dense.

In all the processes of rewriting, the same LLM without any fine-tuning as used in the experiment is employed, proceeding with zero-shot prompting. 
Detailed prompts for each process can be found in Table \ref{table:prompt_details} of Appendix \ref{appendix:details_on_prompts}.

\subsection{Context Rewriting}
In typical retrieval-augmented generation scenarios, the raw context becomes lengthy, making the facts in the context necessary for factuality discrimination sparse. 
To address this issue, context rewriting involves processes of question decomposition and fact extraction followed by fact reorganization.

\textbf{Sub-question Generation} Questions are often complex or require multi-step reasoning. 
Considering this, we first use the LLM to divide the question into multiple sub-questions.
We follow the question decomposition prompt in DDCoT (\citealp{zheng-2023-ddcot}).
Through question decomposition, we aim to facilitate the collection of relevant facts.
For example, for the question \textit{"Are more people today related to Genghis Khan than to Julius Caesar?"}, the generated sub-questions aim to collect facts related to the number of people associated with or descended from Genghis Khan and Julius Caesar, as illustrated in Figure \ref{fig:overview} (a).

\textbf{Question-based Fact Extraction} We then use each sub-question and the context to prompt the LLM to extract the necessary facts from the context. 
Finally, in the fact reorganization process, we prompt the LLM to filter and reorder the extracted facts to construct a filtered context $c'$ using the original question.

\subsection{Query Rewriting}\label{subsec:query_rewriting}
\textbf{Context-based Question Rewriting} In query rewriting, we first rewrite the question $n$ times based on the context $c'$.
The goal of this process is to obtain a statement that is easier for the LLM to understand and ground in the given context.
We denote the rewritten questions including the original one as $Q=\{q_0, q_1, q_2, ..., q_n\}$, where $q_0=q$.
In Figure \ref{fig:overview} (b), for instance, the question is incrementally paraphrased by the LLM into a distribution with a high likelihood, where $n$ is 3.

\textbf{Final Prediction} Using the question $q$, filtered context $c'$, and rewritten questions $Q$, we proceed with conversation-style factuality discrimination.
Initially, we input $q_n$ and $c'$ into the LLM to obtain the prediction $a_n$. 
In the $i$-th step, we input $q_{n-i}$, $c'$, and the dialog history $D_i=\{(q_n, a_n), ..., (q_{n-i+1}, a_{n-i+1})\}$ into the LLM to derive $a_{n-i}$.
Finally, we input the original question $q$ for the statement, along with $c'$ and $D=D_n$, into the LLM to obtain the final factuality prediction.

Figure \ref{fig:overview} assumes the Original prompt setting, and note that in All-True and All-False prompt settings, the final question is converted into statement form.

\section{Details on Prompts}\label{appendix:details_on_prompts}
Table \ref{table:prompt_details} shows the prompts used in the chain-of-question framework. 
We omit the prompt for All-False as they are mostly similar to that for All-True.
Note that we add a sentence "Please return also the reason." to the instruction part of evaluation for examples in Figure \ref{fig:false_negative_problem_intro} and Table \ref{table:additional_examples_of_false_negative_problem} to generate reasoning for predictions.

\section{Prediction Confidence Analysis after Rewriting}\label{appendix:confidence_analysis_rewriting}
We observe changes in prediction confidence when applying context and query rewriting to Mistral. 
We plot the confidence histogram for predicted labels of the counterfactual samples in the All-True and All-False prompt settings, as shown in Figure \ref{fig:confidence_histogram_rewriting}. 

When $n = 0$, only context rewriting is applied to Mistral, the prediction confidence decreases compared to that of baselines. 
As $n$ increases to 5, the number of samples classified as Unknown increases, and among the remaining samples, those classified as True or False show a distribution in higher prediction confidence. 
Assuming that samples classified as Unknown have low confidence in being True or False, the query rewriting stage polarizes the prediction confidence. 
In summary, context rewriting serves to reduce confidence, while query rewriting can be seen as intensifying confidence. 

The difference in confidence distribution between True and False is similar across both prompt types. 
It implies that the confidence of LLMs is influenced not only by the factuality of the statement but also by the content of the response.

\section{License}
The StrategyQA dataset is under the MIT License and the BoolQ dataset is under the cc-by-sa-3.0 License. 
The code of Mistral is under the Apache-2.0 license.

\section{Intended Use of Artifacts}
We use Mistral, ChatGPT, GPT-4, and StrategyQA for academic research purposes and do not used them for improper intentions such as adversarial prompting and jailbreaking.

\section{Computational Cost}
We utilize two Tesla V100 GPUs for inference using Mistral. 
Each inference takes between 2 to 5 minutes.

\section{AI Assistants}
We utilize ChatGPT 4\footnote[2]{\url{https://chat.openai.com/}} for translation, correction, and figure plotting processes.

\end{document}